# Deep Heterogeneous Feature Fusion for Template-Based Face Recognition

Navaneeth Bodla, Jingxiao Zheng, Hongyu Xu, Jun-Cheng Chen, Carlos Castillo, Rama Chellappa

Center for Automation Research, University of Maryland Institute for Advanced Computer Studies
University of Maryland, College Park, MD 20742

{nbodla, jxzheng, hyxu}@umiacs.umd.edu, pullpull@cs.umd.edu, {rama, carlos}@umiacs.umd.edu

## Abstract

*Although deep learning has yielded impressive performance for face recognition, many studies have shown that different networks learn different feature maps: while some networks are more receptive to pose and illumination others appear to capture more local information. Thus, in this work, we propose a deep heterogeneous feature fusion network to exploit the complementary information present in features generated by different deep convolutional neural networks (DCNNs) for template-based face recognition, where a template refers to a set of still face images or video frames from different sources which introduces more blur, pose, illumination and other variations than traditional face datasets. The proposed approach efficiently fuses the discriminative information of different deep features by 1) jointly learning the non-linear high-dimensional projection of the deep features and 2) generating a more discriminative template representation which preserves the inherent geometry of the deep features in the feature space. Experimental results on the IARPA Janus Challenge Set 3 (Janus CS3) dataset demonstrate that the proposed method can effectively improve the recognition performance. In addition, we also present a series of covariate experiments on the face verification task for in-depth qualitative evaluations for the proposed approach.*

## 1. Introduction

Face recognition is one of the active research problems in computer vision. It has several applications including mobile authentication, visual surveillance, social network analysis, and video content analysis. Recently, face recognition systems based on deep convolutional neural networks (DCNNs) [1, 2, 3] have yielded performance surpassing human recognition accuracy on the well-known Labeled Faces in the Wild (LFW) dataset [4, 5]. Although for the LFW dataset, the faces are downloaded from the Internet and thus in unconstrained settings, most of the faces are still near-frontal. For the other well-known video-to-video face dataset such as the Youtube Faces (YTF) dataset [6], although the videos contain more variations in pose and illumination, most of the videos are single-shot and thus contain fewer variations than multiple-shot videos. To develop the next-generation face recognition system, [7] addresses these problems and introduces the IARPA Benchmark A (IJB-A) dataset which contains face images with a full range of pose variations, illumination variations, occlusion, blur, and other challenging conditions. In addition, the dataset uses template-based face identification and verification protocols where a template consists of a set of still face images and video frames from different sources. (*i.e.*, The templates can be (1) image-only templates), (2) video-frame-only templates, or (3) mixed-media templates.). In this paper, we work on the extended version of IJB-A, Janus Challenge Set 3 (Janus CS3), which is a superset of IJB-A and is much more challenging. Janus CS3 contains 1,871 subjects with 11,876 still images and 55,372 video frames sampled from 7,094 videos. Sample templates are shown in Figure 1. Since the templates have lots of variations, the performance of template-based face recognition highly depends on generating a robust feature representation for the templates.

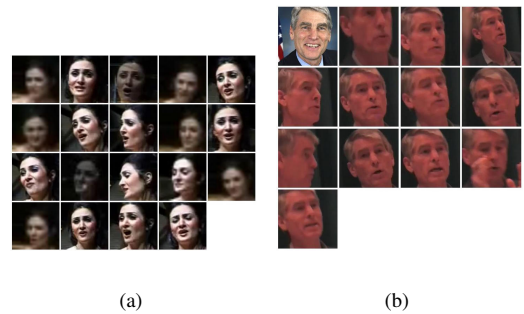

Figure 1. Janus CS3 sample templates which contains a lot of pose, illumination variations, and other challenging conditions.

Over the past five years, deep learning has achieved promising performance in various tasks such as object classification [8], object detection [9], action recognition [10], semantic segmentation [11, 12] and face recognition [1, 2, 3]. With more annotated data and powerful GPUs being readily available, training deeper and larger architectures has become much easier for computer vision tasks. Besides the well-known DCNNs such as Alexnet [13], VGGNet [14], and Resnet [8], there are plenty of variants designed for different tasks and each of them varies with the depth, the number of hidden units, kernel size, activation function, types of connections, loss function, etc. Because these networks are trained for different tasks and objective functions, the features of some networks may be more robust to pose variation than others, and some more robust to illumination changes. There is no one universal network that performs well in all situations, while the top performers for the well-known ImageNet Large Scale Visual Recognition Challenge (ILSVRC) [15] used an ensemble of same or different networks with slightly different settings to achieve top performances. Others such as Ranjan *et al.* in [16] used a single network to train an ensemble of different face related tasks in a multitask learning framework that boosts the performance of individual tasks. Hence, there is a need of ensemble learning to enhance the performance, be it an ensemble of multiple tasks [16], or multiple networks fused together. We adopt the latter approach for performing enhanced feature-level fusion of deep features to exploit the strengths of different deep architectures

In order to obtain an improved feature representation, we address two issues: 1) lack of one universal deep neural network and 2) challenges in template-based recognition systems by proposing a feature-level fusion method for heterogeneous deep features from two complementary deep networks. The main contributions are summarized as follows:

- We propose a novel deep fusion network that fuses two state-of-the-art deep features for template-based face recognition to produce a more discriminative representation.

- While performing fusion of the features in a template, the proposed method also takes the inherent geometry of these features into consideration to produce an enhanced template-based representation.

- In addition to traditional quantitative evaluations using the ROC curves, we present a series of qualitative analysis of covariates for the face verification task on the Janus CS3 dataset.

The rest of this paper is organized as follows: in Section 2, we present a brief review of related work and traditional feature fusion methods. In Section 3, we discuss the proposed approach for fusing two CNN features to generate discriminative template-based feature representation. In Section 4, we present the experimental results and an analysis of the face verification task on Janus CS3 dataset. Finally, we conclude the paper in Section 5.

## 2. Related Work

In this section, due to a large amount of related papers, we briefly review some relevant fusion methods drawn from two main categories: (1) traditional feature fusion methods and (2) neural network-based feature fusion methods.

### 2.1. Traditional Feature Fusion

Multiple Kernel Learning (MKL) is one of the most widely used feature-fusion methods. The key idea behind MKL is to combine multiple kernels in an efficient way as one single kernel to maximize the desired prediction accuracy. The multiple kernels could utilize features from different modalities, different feature extraction techniques or features with different notions of similarities [17]. Researchers have proposed a number of variations of MKL by imposing various sparsity conditions over the kernel coefficients. In [18], Yeh *et al.* proposed a Group Lasso regularized MKL for heterogeneous feature fusion. Canonical Correlation Analysis (CCA) is a joint dimensionality reduction/subspace learning method where the two features of same or different dimensionality are projected onto a common subspace such that the correlation of the projected vectors is maximized. The details of CCA can be found in [19]. In [20], Fu *et al.* proposed a method to find a general linear subspace in which cumulative pairwise canonical correlation is maximized after dimensionality reduction and subspace projection.

### 2.2. Neural Network-Based Feature Fusion

Various feature fusion methods have been proposed based on deep neural networks but are mainly for video classification tasks such as action recognition to combine spatial-appearance and temporal streams [21, 22, 23]. For spatial-appearance feature fusion, bilinear CNNs (BCNNs) have been proposed recently for fusing two DCNN features designed for visual recognition [24]. Fusion is achieved through computing the outer product of last-layer features for the two streams of DCNNs followed with softmax loss for model training. In [25], Chowdhury *et al.* used BCNNs for face recognition and demonstrated good results. (*i.e.*, both streams use the VGGNet [14] architectures with different depths respectively.) They also point out that the final features act as applying a polynomial kernel to input features. A multiplicative feature fusion has been proposed in [26] where the feature fusion of two networks is acomplished through element-wise multiplication with bias. Au-toencoders have also been researched for unsupervised fusing of information [27].

The proposed fusion method is different from the methods discussed above as we focus on feature fusion of DC-NNs with different architectures and different feature dimensionality. Unlike BCNN and other methods for video classification where the features to be fused are of same dimensionality and are extracted from two networks with similar architectures the heterogeneity of our method comes from fusion of two different deep CNNs which capture complementary features.

## 3. Proposed Method

The proposed method for fusion of two deep features is illustrated in Figure 2b. Network I [28] and network II [29] are the two pre-trained DCNNs that produce two features called raw features ($\mathbf{x} \in \mathbb{R}^{d_1 \times 1}$ and $\mathbf{y} \in \mathbb{R}^{d_2 \times 1}$ where $d_1$ and $d_2$ are the dimensions of features produced by networks I and II). Network I takes as input faces from tight bounding boxes as input while network II takes faces from loose bounding boxes as input and thus contains more context information from background. We fuse these two complementary raw features by projecting them onto non-linear high-dimensional spaces using a deep neural network as detailed below. Our deep fusion network produces a fused feature vector $\mathbf{z} \in \mathbb{R}^{d \times 1}$ where $d$ is the dimensionality of the fused feature vector.

### 3.1. Basic Building Block of Fusion Network

We illustrate the proposed approach with a simple three-layer network as shown in Figure 3. The hidden layer in this example has more activation units than input and is followed by a non-linear activation function, like tanh. For this network, the inputs are raw features, $\mathbf{x}$, extracted from a pre-trained DCNN, then the output of the hidden layer after non-linear activation function is a non-linear high-dimensional projection, $\phi(\mathbf{x})$, of the input $\mathbf{x}$. If $\mathbf{x}$ and $\phi(\mathbf{x})$ are concatenated, the dual form of the concatenated feature vector $\mathbf{z}$ (where $\mathbf{z}^\mathsf{T} = \begin{bmatrix} \mathbf{x}_i^\mathsf{T} & \phi(\mathbf{x}_i^\mathsf{T}) \end{bmatrix}$) corresponds to a combination of kernels as shown below in (1).

$$\begin{aligned} k(\mathbf{z}_i, \mathbf{z}_j) &= \mathbf{z}_i^\mathsf{T} \mathbf{z}_j \\ &= \begin{bmatrix} \mathbf{x}_i^\mathsf{T} & \phi(\mathbf{x}_i^\mathsf{T}) \end{bmatrix} \begin{bmatrix} \mathbf{x}_j \\ \phi(\mathbf{x}_j) \end{bmatrix} \\ &= k_1(\mathbf{x}_i, \mathbf{x}_j) + k_2(\phi(\mathbf{x}_i), \phi(\mathbf{x}_j)) \end{aligned} \quad (1)$$

where $k_1(\mathbf{x}_i, \mathbf{x}_j) = \mathbf{x}_i^\mathsf{T} \mathbf{x}_j$ is the kernel corresponding to the raw features, $k_2(\phi(\mathbf{x}_i), \phi(\mathbf{x}_j)) = \phi(\mathbf{x}_i)^\mathsf{T} \phi(\mathbf{x}_j)$ is the kernel corresponding to the non-linear high-dimensional projection of raw features and $k(\mathbf{z}_i, \mathbf{z}_j)$ is the combination of these two kernels. Then this three layer network along with a concatenation connection can be trained using hinge loss, and the resulting features extracted from the concatenated layer can be used for the classification tasks such as face identification or can further be followed by metric learning (*e.g.*, Joint Bayesian metric learning [30, 28] or triplet probabilistic embedding [29]) for the face verification task. Figure 3 illustrates the concatenation connection for a simple three layer network. Though in this example we used a single hidden layer, more than one hidden layers can be used to take advantage of deeper architectures of neural networks. We summarize this idea as basic building block of our fusion network in Figure 2a. The basic fusion block takes a raw DCNN feature as input and produces a concatenated feature after a series of non-linear high-dimensional projections implemented by fully connected layers followed by non-linear activation functions.

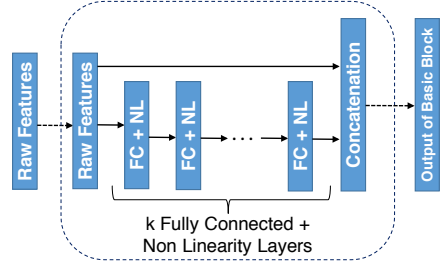

(a) Basic block of fusion network

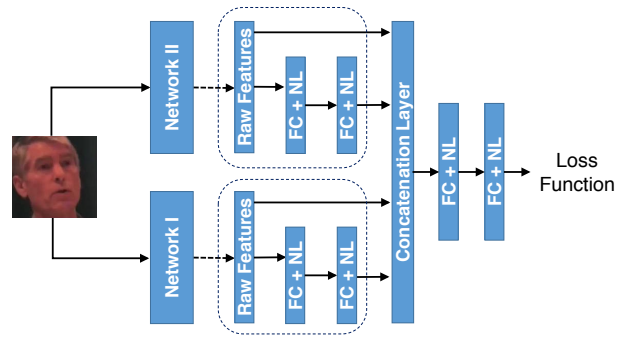

(b) Deep fusion network

Figure 2. The figure on top shows our basic building block of deep fusion network with $k$ hidden layers. The bottom figure is the overview of our deep fusion network that fuses two DCNN features. The boundaries shown in dotted lines are the two basic blocks corresponding to deep networks I [28] and II [29].

### 3.2. Heterogeneous Fusion of Deep Features

For heterogeneous fusion of deep features, each raw DCNN feature first passes through a basic block to generate a concatenated feature vector. The outputs of the basic blocks are further concatenated to generate a high-dimensional feature vector. In general, if there are $k$ fully connected layers per basic block, then all the $k$ outputs of the fully connected layers can be concatenated by treating each of them as a high-dimensional projection. $k$ here is a hyperparameter and in our case we set it to 1, that is we concatenate only the output of last fully connected layer of each basic block. If there are $m$ input sources that generate raw features and each implements a basic block with one additional high-dimensional projection from the last fully connected layer, then we get a total of $2m$ concatenations in the final feature vector. The fusion network is shown in Figure 2. In this paper, we fuse features from two networks, that is set $m$ to 2 and concatenate the corresponding four features into a higher-dimensional representation as shown in (2)

$$\mathbf{z}^\mathsf{T} = \begin{bmatrix} \mathbf{x}^\mathsf{T} & \mathbf{y}^\mathsf{T} & \phi_1(\mathbf{x}^\mathsf{T}) & \phi_2(\mathbf{y}^\mathsf{T}) \end{bmatrix} \quad (2)$$

where $\mathbf{x}$ and $\mathbf{y}$ are the raw features of DCNN I and DCNN II respectively. $\phi_1(\mathbf{x}^\mathsf{T})$ is the non-linear high-dimensional projection of $\mathbf{x}$ and $\phi_2(\mathbf{y}^\mathsf{T})$ is the non-linear high-dimensional projection of $\mathbf{y}$. Since each feature has its own scale, we first $l_2$ normalize each of the feature vector before concatenation.

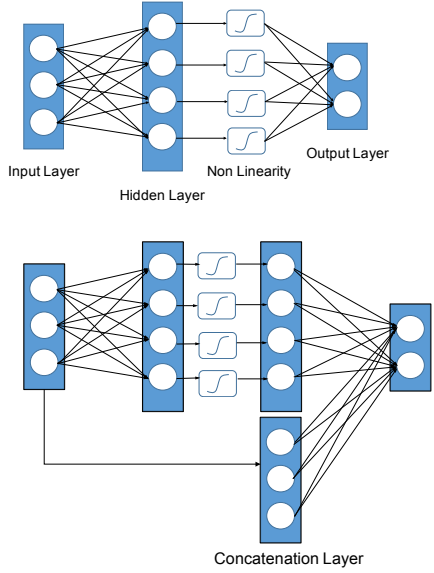

Figure 3. An example of a three-layer network illustrating our approach. The figure on the top shows a simple three layer network and the figure on the bottom shows how to convert it to basic block of fusion network by concatenating input with output of hidden layer after non-linear activation functions.

| Input I Size: 512 | | Input II Size: 320 | |
|---|---|---|---|
| **fc$_I$1** : FC in basic block 1 | | **fc$_{II}$1** : FC in basic block 2 | |
| Output Size | 1,024 | Output Size | 512 |
| # Params | 0.52M | # Params | 0.16M |
| **fc$_I$2** : FC in basic block 1 | | **fc$_{II}$2** : FC in basic block 2 | |
| Output Size | 2,048 | Output Size | 1,024 |
| # Params | 2.09M | # Params | 0.52M |
| **Concatenation Layer** | | Output Size: 3,904 | |
| Name | Descripton | Output Size | # Params |
| **fc1** | Fully Connected | 4,906 | 19.1M |
| **fc2** | Fully Connected | 2,048 | 10.04M |
| **cost** | softmax | 1,710 | 3.5M |
| total | | | 35.9M |

Table 1. Details of our deep fusion network.

Our network when trained with hinge loss on concatenated feature layer resembles Multiple Kernel Learning methods. However the main difference and the advantage of our network is that in traditional MKL methods, the base kernels are handpicked after cross validation but in our case the high dimensional projections $\phi_1$ and $\phi_2$ are jointly learned via back propagation during training. Another advantage of this approach is that the network produces features, which can be further used to perform metric learning and computing similarity for face verification which is not possible in traditional methods where MKL is solved in the dual space. Because the high-dimensional projections are learned jointly, our network is able to learn more discriminative representation by combining information from both the networks. In our implementation, we consider two versions of our fusion network: 1) network trained with hinge loss (HL) directly on top of concatenation layer and 2) network trained with two additional fully connected layers on top of the concatenation layer and trained with softmax loss (SL). In both cases, the networks are trained using a standarad stochastic gradient descent algorithm using caffe [31] and the final fused features are extracted from concatenation layer. The details of the two deep networks that generate raw features can be found in [28] and [29], and the architecture of the fusion network is shown in table 1.

### 3.3. Template Representative Fused Feature

Another key aspect of our implementation is how to generate template representative fused feature. Since each template has a variable number of faces from still images or video frames, before performing template-based face recognition, we have to aggregate all the features together to a compact feature representation. Average pooling is the most common operation for this purpose. However, average pooling is not applicable to our fused features and deteriorates the performance because it ignores the inherent geometry of the features.

Since the final fused vector is learned through a deep neural network, it can be represented by a function of input features $\mathbf{x}$ and $\mathbf{y}$ as $\mathbf{z} = f(\mathbf{x}, \mathbf{y})$ where $f$ is a non-linear function (*i.e.*, $f$ refers to the fusion network in this paper). Consider a template with $n$ faces (images plus frames). Let $\mathbf{x}_1, \ldots, \mathbf{x}_n$ be the raw feature vectors corresponding to network I and $\mathbf{y}_1, \ldots, \mathbf{y}_n$ be the features corresponding to network II. The corresponding fused feature be $\mathbf{z}_i$ that is, $\mathbf{z}_1 = f(\mathbf{x}_1, \mathbf{y}_1), \ldots, \mathbf{z}_n = f(\mathbf{x}_n, \mathbf{y}_n)$. The traditional average pooled features $\mathbf{z}_i$'s is given in (3):

$$\hat{\mathbf{z}} = \sum_{i=1}^{n} \lambda_i \mathbf{z}_i \quad (3)$$

where $\lambda_i$ is a weight scalar (i.e. $\lambda_i = \frac{1}{n}$.). $\hat{\mathbf{z}}$ is computed by the weighted average of fused features in Euclidean space. Therefore, $\hat{\mathbf{z}}$ may not lie in the non-linear space formed by the fusion network. In order to handle this problem, and preserve the inherent geometry of the fused features, we first do the average pooling of the raw features and then fuse the averaged raw features as shown in (4). We use the average pooled features $\tilde{\mathbf{x}}$ and $\tilde{\mathbf{y}}$ as inputs to our fusion network to yield the template representative fused feature vector $\tilde{\mathbf{z}}$. An illustration of fused feature space is shown in Figure 4.

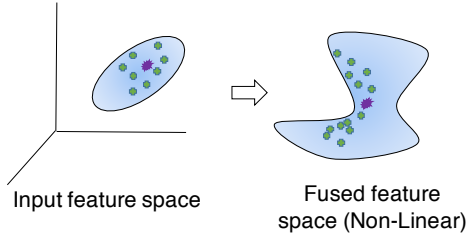

Figure 4. Illustration of fused features lying on a non-linear space.

$$\tilde{\mathbf{x}} = \sum_{i=1}^{n} \lambda_i \mathbf{x}_i, \tilde{\mathbf{y}} = \sum_{i=1}^{n} \lambda_i \mathbf{y}_i, \tilde{\mathbf{z}} = f(\tilde{\mathbf{x}}, \tilde{\mathbf{y}}) \quad (4)$$

For evaluating the similarity between two templates A and B, we compute the cosine distance between the respective template representative fused features $\tilde{\mathbf{z}}_A$ and $\tilde{\mathbf{z}}_B$.

## 4. Experiments

In this section we present the experimental results for template-based face recognition on IARPA's Janus Challenge Set 3 (Janus CS3). The dataset is introduced first and then followed by the evaluation results and discussions.

### 4.1. IARPA Janus Challenge Set 3

To the best of our knowledge, IARPA Janus Benchmark A (IJB-A) [7] is the first template-based face recognition dataset which contains the faces of 500 different subjects with full pose, large illumination variations and other extreme conditions. However, for this work, we focus on its extended version, IARPA's Janus Challenge Set 3 (Janus CS3), which is a superset of the IJB-A dataset. It has 1,870 subjects with 11,876 still image and 55,372 video frames sampled from 7,094 videos. There are three different protocol settings for face verification and identification and each protocol is evaluated with a subset of the CS3 dataset. The three protocols are respectively: 1) 1:1 template-based face verification (*i.e.*, templates can have both still images or frames of a video with varying number of images in each template), 2) open set 1:N protocol for face identification for image-only probe templates and open set 1:N protocol for face recognition with mixed-media probe templates (both images and video frames), and 3) 1:1 face verification with covariate analysis. A few samples of faces in templates are shown in Figure 1. In addition, different from IJB-A, the Janus CS3 dataset does not provide any training data. Therefore, we use a subset of UMDFace [32] as the training dataset which consists of with 89,177 face images of 1,710 unique subjects from the Internet to train our deep fusion network. There are no overlapping subjects between the collected dataset and the Janus CS3 dataset.

### 4.2. Implementation Details

In this section, we provide the implementation details of our approach. We evaluate our deep fusion network with deep features of two state-of-the-art template-based face recognition methods as in [29] and [28]. The deep features are first extracted using the DCNN architectures presented in in [28, 29] for our training dataset and our network shown in Figure 2 is trained using the standard stochastic gradient descent method with a learning rate of 0.001, batch size of 1,000 for 2,000 iterations. The two different heterogeneous features are of dimensions 512 and 320 and the learned fused feature dimension is 3,904. We further reduce the feature dimension to 140 by Principle Component Analysis (PCA). The similarity score is computed based on the cosine similarity distance [33]. In addition, we also perform the joint Bayesian metric learning as in [28] to further evaluate the effectiveness of our approach.

**Methods for Comparison**: We first evaluate our features, learned by fusion network on template-based face verification without the metric learning techniques and compare it with several state-of-the-art architectures listed below:

- Baseline 1 [29]. We compute the cosine similarity score [33] directly from the features learned in [29].

- Baseline 2 [28]. We compute the cosine similarity score [33] directly from the features learned in [28]

- Baseline 1+2. Score level fusion by averaging the two similarity scores from baseline 1 and 2.

- Bi-auto [27]. The two features are fused in an unsupervised way using a bimodal autoencoder as in [27]. We compute the cosine similarity score [33] of the shared representation features.

- Concat [26]. The two input features are concatenated followed by two fully connected layers as in [26]. Cosine similarity score [33] is then computed using the last fully connected layer features.

We report our results on both (1) the hinge loss on top of concatenation layer (**Ours(HL)**) and (2) softmax loss with two additional fully connected layers on top of the concatenation layer (**Ours(SL)**). In order to further demonstrate the effectiveness of our approach, we perform metric learning [28, 29] on the two deep fusion features and all the other compared deep CNN features.

### 4.3. Template-based Face Verification

We first evaluate the proposed fusion network for the template-based face verification protocol of Janus CS3 dataset. In this protocol, a probe template needs to be verified against a gallery template to check if it belongs to the same person or not. The Janus CS3 dataset has over 7.8 million comparisons for this protocol using mixed-media templates (images and video frames). It has about $1,870$ unique subjects with a total of $12,590$ templates. There are $10,694$ genuine and $7,792,066$ impostor matches. Each template contains a variable number of faces from still images and video frames. The gallery templates contain only images but the probe templates contain both images and frames.

The template-based feature representation for a template is first extracted as described in Section 3.3 for fused features followed by cosine similarity distance [33] or learned similarity measure with metric learning to compute the similarity scores. In contrast, the template-based feature representation for the baseline algorithm is computed by aggregating individual features using average pooling. We compare our results with (1) baseline 1 [29], (2) baseline 2 [28], (3) baseline $1+2$, (4) Bi-auto [27] and (5) Concat [26] and plot the ROC curves in Figure 5. We also report the various true positive rates for false positive rates at $1e-4$, $1e-3$, $1e-2$ and $1e-1$ in Table 2. First, our fusion method outperform baseline $1+2$ significantly, which is the score level fusion of the DCNN features in [29] and [28]. This indicates that the proposed deep fusion network is able to exploit the correlations between the two heterogeneous features to produce more discriminative fused features. Second, from the bottom half of the Table 2, it can be seen that metric learning techniques could improve the verification performance for all the methods. However, our methods **Ours(HL)** and **Ours(SL)** still consistently outperform all the compared methods by a large margin, which demonstrates the effectiveness of the proposed fusion network. Figure 6 shows a few example gallery and probe templates where our fusion method performs better than baseline $1+2$.

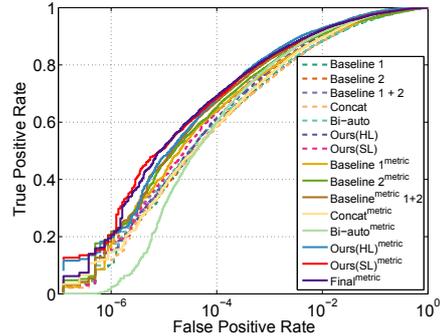

Figure 5. CS3 1:1 verification protocol

| Face Verification | 1e-04 | 1e-03 | 1e-02 | 1e-01 |
|---|---|---|---|---|
| Baseline 1 [29] | 0.5877 | 0.7686 | 0.8986 | 0.9690 |
| Baseline 2 [28] | 0.5813 | 0.7544 | 0.8834 | 0.9616 |
| Baseline 1 + 2 | 0.6051 | 0.7819 | 0.9033 | 0.9700 |
| Bi-auto [27] | 0.6047 | 0.7630 | 0.8849 | 0.9623 |
| Concat [26] | 0.5831 | 0.7476 | 0.8823 | 0.9716 |
| Ours (HL) | *0.6148* | *0.7823* | *0.9032* | 0.9699 |
| Ours (SL) | **0.6453** | **0.8155** | **0.9242** | **0.9842** |
| With Metric Learning | 1e-04 | 1e-03 | 1e-02 | 1e-01 |
| Baseline 1$^{metric}$ [29] | 0.6515 | 0.8000 | 0.9082 | 0.9674 |
| Baseline 2$^{metric}$ [28] | 0.6406 | 0.7845 | 0.8979 | 0.9674 |
| Bi-auto$^{metric}$ [27] | 0.6010 | 0.7969 | 0.9148 | 9761 |
| Concat$^{metric}$ [26] | 0.5877 | 0.7647 | 0.8960 | 0.9755 |
| Ours (HL)$^{metric}$ | 0.6808 | **0.8364** | **0.9291** | **0.9856** |
| Ours (SL)$^{metric}$ | *0.6850* | 0.8271 | 0.9222 | *0.9786* |
| Final 1 | **0.6981** | 0.8289 | 0.9229 | 0.9774 |
| Final 2 | 0.6845 | *0.8297* | *0.9245* | 0.9781 |

Table 2. Verification performance of different DCNN features for the Janus CS3 dataset. The true positive rates (TPR) at false positive rate (FPR) of 0.0001, 0.001, 0.01 and 0.1 are reported. In the upper half of the table, we present the results before metric learning. In the bottom half of the table, the results are shown after metric learning. Final 1 refers to (Baseline 1 + Baseline 2 + Ours(SL))$^{metric}$, and Final 2 is (Baseline 1 + Baseline 2 + Ours(HL))$^{metric}$.

### 4.4. Open Set Face Identification

We further evaluate the proposed approach for the template-based face identification protocol of the Janus CS 3 dataset where the identity of the person from the probe needs to be identified by comparing with the enrolled subjects a gallery. This is $1:N$ search protocol where $N$ is the total number of subjects in the gallery. Unlike traditional closed-set face identification, the Janus CS3 dataset also focuses on the open set setting where the subject from the probe may not be matched to any subjects in the gallery.

There are two different template settings in the protocol of open set identification. (1) For setting 1, both gallery and probe consist of image-only templates. (2) For setting 2, it differs from setting 1 in that the probe templates contain faces from both images and video frames. In each setting, the gallery data is further divided into two non-overlapping splits. Split 1 in the gallery contains $940$ subjects and split 2

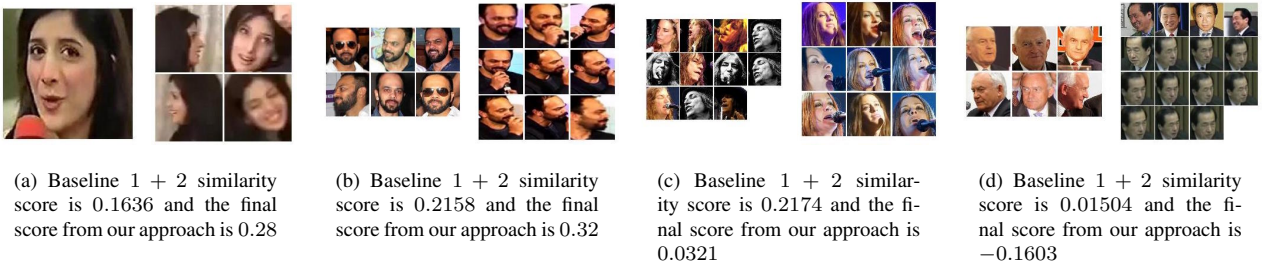

(a) Baseline $1+2$ similarity score is 0.1636 and the final score from our approach is 0.28

(b) Baseline $1+2$ similarity score is 0.2158 and the final score from our approach is 0.32

(c) Baseline $1+2$ similarity score is 0.2174 and the final score from our approach is 0.0321

(d) Baseline $1+2$ similarity score is 0.01504 and the final score from our approach is $-0.1603$

Figure 6. Similarity scores of two example pair of templates from Baseline $1+2$ and our approach. Note that for the pair of templates from the same subject (fig 6a and b), our fusion network generates higher similarity score than Baseline $1+2$ and for the pair of templates from different subjects (fig 6c and d), our fusion network generate lower similarity scores compared to Baseline $1+2$.

contains the rest of the 930 subjects. Probe data in each settings contain all $1,870$ subjects. However, in setting 1, the template from probe is only image-based (probe-img) while the probe template is mixed media-based (probe-mixed)in setting 2.

We compute the average rank 1/5/10 accuracies to evaluate the closed-set identification performance and true positive identification rate vs false positive identification rate [34] (**TPIR/FPIR**) curves for the open-set identification performance from all the methods. The similarity score between probe template and gallery template is computed by learned similarity measurement with metric learning [28]. We also compare the proposed method with traditional MKL-SVM [35]. We train the one-vs-all SVMs for all the enrolled subjects in the gallery with linear and radial basis functions for both the DCNN features where the bandwidth ($\gamma = 0.0625$) of the kernel is chosen by cross-validation and the cost (C) is chosen as a ratio of number of negative templates to positive templates. Gallery and probe templates are used for training and testing respectively whereas the probability estimates computed by the MKL-SVM are used as similarity scores.

It can be seen from the TPIR/FPIR curves that for the split 2 in setting 1 (gallery S2 vs probe-image) and setting 2 ( gallery S2 vs probe-mixed), our deep fused features consistently outperform all the other methods. Our approach also achieves the best performance for split 1 in both settings (gallery S1 vs probe-image and gallery S1 vs probe-mixed). It is also noted that the open-set identification task is more challenging than close-set identification task. We also sum the three scores up (Baseline 1, 2 and Ours(SL)) and report the final performance in Table 3. The plots of rank/accuracy are given in the supplementary material.

### 4.5. Face Verification with Covariate Analysis

We further evaluate the proposed method on Janus CS3 face verification with covariate analysis. This evaluation protocol is unique to Janus CS3, which enables a qualitative analysis of the relevance of eight different covariates

| Setting 1 | Methods | Rank Accuracy | | | TPIR | |
|---|---|---|---|---|---|---|
| | | Rank 1 | Rank 5 | Rank 10 | FPIR@1e-1 | FPIR@1e-2 |
| Split 1 | Baseline $1+2$ | 0.817 | 0.899 | 0.923 | 0.719 | 0.473 |
| | Bi-auto [27] | 0.789 | 0.880 | 0.903 | 0.688 | 0.406 |
| | Concat [26] | 0.754 | 0.850 | 0.884 | 0.620 | 0.346 |
| | MKL [35] | 0.664 | 0.765 | 0.824 | 0.585 | 0.434 |
| | Ours(HL) | 0.814 | 0.893 | 0.920 | 0.724 | 0.452 |
| | Ours(SL) | 0.812 | 0.897 | 0.921 | 0.719 | **0.507** |
| | Final | **0.821** | **0.903** | **0.926** | **0.728** | 0.484 |
| Split 2 | Baseline $1+2$ | 0.826 | 0.904 | 0.927 | 0.735 | 0.500 |
| | Bi-auto [27] | 0.802 | 0.889 | 0.913 | 0.706 | 0.385 |
| | Concat [26] | 0.751 | 0.864 | 0.889 | 0.640 | 0.408 |
| | MKL [35] | 0.698 | 0.783 | 0.836 | 0.617 | 0.485 |
| | Ours(HL) | 0.813 | 0.900 | 0.922 | 0.734 | 0.427 |
| | Ours(SL) | 0.816 | 0.905 | 0.929 | 0.739 | 0.534 |
| | Final | **0.828** | **0.909** | **0.929** | **0.747** | **0.542** |
| Setting 2 | Methods | Rank Accuracy | | | TPIR | |
| | | Rank 1 | Rank 5 | Rank 10 | FPIR@1e-1 | FPIR@1e-2 |
| Split 1 | Baseline $1+2$ | 0.853 | 0.917 | 0.935 | 0.758 | 0.484 |
| | Bi-auto [27] | 0.827 | 0.901 | 0.924 | 0.708 | 0.430 |
| | Concat [26] | 0.791 | 0.879 | 0.913 | 0.649 | 0.381 |
| | MKL [35] | 0.600 | 0.687 | 0.757 | 0.541 | 0.385 |
| | Ours(HL) | 0.849 | 0.917 | 0.935 | 0.750 | 0.474 |
| | Ours(SL) | 0.842 | 0.917 | 0.936 | 0.755 | **0.530** |
| | Final | **0.854** | **0.922** | **0.938** | **0.771** | 0.517 |
| Split 2 | Baseline $1+2$ | 0.812 | **0.897** | 0.917 | 0.727 | 0.499 |
| | Bi-auto [27] | 0.768 | 0.862 | 0.897 | 0.705 | 0.459 |
| | Concat [26] | 0.742 | 0.848 | 0.883 | 0.656 | 0.418 |
| | MKL [35] | 0.597 | 0.686 | 0.742 | 0.521 | 0.413 |
| | Ours(HL) | 0.795 | 0.884 | 0.914 | 0.751 | 0.490 |
| | Ours(SL) | 0.800 | 0.889 | 0.914 | 0.745 | 0.526 |
| | Final | **0.810** | 0.895 | **0.920** | **0.744** | **0.537** |

Table 3. Open Set Face Identification. Average rank 1/5/10 accuracies and TPIR at FPIR $= 0.1, 0.01$ are reported. All the results are reported after metric learning [28].

(*i.e.* age, eyes visible, facial hair, forehead visible, gender, nose and mouth visible, indoor and skin tone) in face verification. The evaluation of the covariate verification protocol is over 20 million $1:1$ comparison using single image-based templates. The test set contains $20,866,895$ pair of templates (5,961,839 genuine and 14,905,056 poster pairs).

This evaluation provides insights into which covariate the features are robust to and what they are good at representing. More specifically, given a pair of templates, a particular covariate can be presented in both the templates have such covariate or not. Taking 'eyes visible'as example, two comparing templates could both have the eyes visible, or only one is eyes visible, or both of them are eyes invisible.

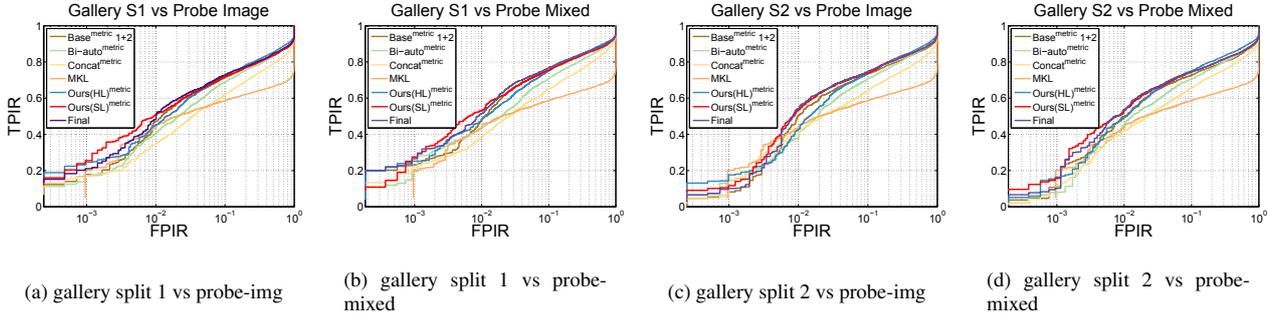

(a) gallery split 1 vs probe-img   (b) gallery split 1 vs probe-mixed   (c) gallery split 2 vs probe-img   (d) gallery split 2 vs probe-mixed

Figure 7. Janus CS3 search protocol: FPIR vs TPIR for Evaluation 2, tasks 1 and 2.

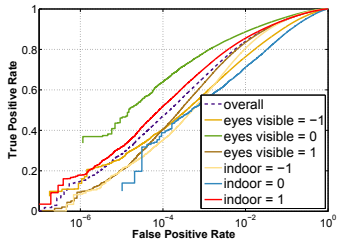

Figure 8. Janus CS3 covariate protocol overall verification performance.

|  | 1e-04 | 1e-03 | 1e-02 | 1e-01 |
|---|---|---|---|---|
| Overall | 0.4707 | 0.6639 | 0.8396 | 0.9488 |
| eyes visible $= -1$ | 0.4044 | 0.5797 | 0.7636 | 0.9143 |
| eyes visible $= 0$ | 0.6392 | 0.7846 | 0.8868 | 0.9605 |
| eyes visible $= 1$ | 0.4048 | 0.6343 | 0.8346 | 0.9480 |
| indoor $= -1$ | 0.3508 | 0.5661 | 0.8052 | 0.9488 |
| indoor $= 0$ | 0.3884 | 0.5450 | 0.7114 | 0.8848 |
| indoor $= 1$ | 0.5188 | 0.7055 | 0.8552 | 0.9494 |

Table 4. Covariate (indoor, eye visible) and overall verification performance for the Janus CS3 covariate protocol.

If the the pair of templates have the covariate in common, we use a number $> -1$ to describe the covariate. Otherwise, we set the number $= -1$ for the covariate, which indicates that the pair of templates do not have the covariate in common. We perform two series of experiments to evaluate all the test pairs in CS3 covariate protocol:

**1. Overall Verification Performance**. We plot the overall ROC curve in Figure 8 without specifying the covariates. This provides a baseline for all 8 covariates to compare with. In addition, we report the average TPR at various FPR in Table 4.

**2. Different Covariates Performance**: In order to evaluate the effect of different covariates on the verification performance, we plot separate ROC curves for 2 covariates (indoor and eyes visible) in Figure 8 as an example. From the ROC curve in Figure 8 we observe that the verification performance is better when the the pair of templates are both taken indoors compared to the other pair of templates, which at least one of them is outdoor. This shows that the learned deep features are more robust to the same indoor lighting condition. Moreover, for the covariate 'eyes visible' shown in Figure 8, it can be seen that having the eyes visible in both two templates gives considerable improvement in verification performance. The analysis of ROC curves for the eight covariates are given in the supplementary material.

## 5. Conclusion

In this paper, we presented a novel deep heterogeneous feature fusion network approach for template-based face recognition. The proposed approach efficiently fused the discriminative information of different deep features by 1) jointly learning the non-linear high-dimensional projection of the deep features and 2) generating a more discriminative template feature representation, which preserves the inherent geometry of deep features in the feature space. We extensively evaluated our approach on the IARPA Janus Challenge Set 3 (Janus CS3) for both template-based face verification and identification, in addition, we also presented qualitative evaluations of face verification performance with covariate analysis. The experimental results clearly demonstrate the effectiveness of the proposed approach and the improvements over the state-of-the-art.

## 6. Acknowledgement

This research is based upon work supported by the Office of the Director of National Intelligence (ODNI), Intelligence Advanced Research Projects Activity (IARPA), via IARPA R&D Contract No. 2014-14071600012. The views and conclusions contained herein are those of the authors and should not be interpreted as necessarily representing the official policies or endorsements, either expressed or implied, of the ODNI, IARPA, or the U.S. Government. The U.S. Government is authorized to reproduce and distribute reprints for Governmental purposes notwithstanding any copyright annotation thereon.

# Deep Heterogeneous Feature Fusion for Template-Based Face Recognition
## Supplementary Material

Navaneeth Bodla    Jingxiao Zheng    Hongyu Xu    Jun-Cheng Chen
Carlos Castillo    Rama Chellappa
Center for Automation Research, UMIACS, University of Maryland, College Park, MD 20742
{nbodla, jxzheng, hyxu}@umiacs.umd.edu, pullpull@cs.umd.edu, {rama, carlos}@umiacs.umd.edu

**Abstract**

*In this supplementary material, we provide more details on the two plots mentioned in the main paper: (1) The plots of rank accuracy of open set face identification. (2) The analysis and ROC curves of the eight covariates in face verification with covariates analysis protocol.*

## 1. Open Set Face Identification

In this section we plot rank vs accuracy curves for face identification on Janus CS3 dataset. We compute and plot the accuracy for top 1 to 500 ranks for our and all the compared methods as shown in figure 1.

## 2. Face Verification with Covariate Analysis

In this section, we plot eight separate ROC curves for all the test pairs over eight covariates (*i.e.* age, eyes visible, facial hair, forehead visible, gender, nose and mouth visible, indoor and skin tone). We also provide the overall verification performance as a baseline for all 8 covariates to compare with. In addition, we report the average true positive rate (TPR) at false positive rate (FPR) = $1e-4$, $1e-3$, $1e-2$ and $1e-1$ from Table 1 to Table 8.

**Age**: Age $= -1$ means for the compared pair, the age of the person in one image is not the same as the other one. Age $> -1$ means that the age of the person are the same for both images.

**Eyes visible**: Eyes visible $= -1$ indicates that for the compared pair, eyes are visible in one image and invisible in the other one. Eyes visible $= 0$ means eyes are invisible for both images and eyes visible $= 1$ means eyes are visible for both images. The performance for eyes visible $= 0$ is better than the other two cases.

**Facial hair**: Facial hair $= -1$ means for the compared pair, the amount of facial hair in one image is not the same as the other one. Facial hair $= 0$ means facial hair does not exist for both images and Facial hair $> 0$ means that the amount of facial hair are of the same amount for both images.

**Forehead visible**: Forehead visible $= -1$ means for the compared pair, forehead are visible in one image and invisible in the other one. Forehead visible $= 0$ means forehead are invisible for both images and forehead visible $= 1$ means forehead are visible for both images. The performance for forehead visible $= 0$ is worse than the other two cases.

|            | 1e-04  | 1e-03  | 1e-02  | 1e-01  |
|------------|--------|--------|--------|--------|
| Overall    | 0.4707 | 0.6639 | 0.8396 | 0.9488 |
| age = −1   | 0.4740 | 0.6716 | 0.8490 | 0.9533 |
| age = 0    | 0.5283 | 0.6650 | 0.8383 | 0.9480 |
| age = 1    | N/A    | 0.3174 | 0.4384 | 0.7279 |
| age = 2    | 0.3166 | 0.5468 | 0.7624 | 0.8860 |
| age = 3    | 0.5816 | 0.7511 | 0.8650 | 0.9433 |
| age = 4    | 0.4501 | 0.6177 | 0.8120 | 0.9459 |
| age = 5    | 0.3753 | 0.5351 | 0.7554 | 0.9221 |

Table 1. Covariate "age" verification performance for the Janus CS3 covariate protocol.

|                   | 1e-04  | 1e-03  | 1e-02  | 1e-01  |
|-------------------|--------|--------|--------|--------|
| Overall           | 0.4707 | 0.6639 | 0.8396 | 0.9488 |
| eyes visible = −1 | 0.4044 | 0.5797 | 0.7636 | 0.9143 |
| eyes visible = 0  | 0.6392 | 0.7846 | 0.8868 | 0.9605 |
| eyes visible = 1  | 0.4048 | 0.6343 | 0.8346 | 0.9480 |

Table 2. Covariate "eyes visible" verification performance for the Janus CS3 covariate protocol.

|                  | 1e-04  | 1e-03  | 1e-02  | 1e-01  |
|------------------|--------|--------|--------|--------|
| Overall          | 0.4707 | 0.6639 | 0.8396 | 0.9488 |
| facial hair = −1 | 0.5086 | 0.6338 | 0.7745 | 0.9154 |
| facial hair = 0  | 0.4545 | 0.6580 | 0.8426 | 0.9513 |
| facial hair = 1  | 0.5884 | 0.7784 | 0.8844 | 0.9499 |
| facial hair = 2  | 0.6890 | 0.7830 | 0.9012 | 0.9754 |
| facial hair = 3  | 0.4059 | 0.6076 | 0.7620 | 0.8994 |

Table 3. Covariate "facial hair" verification performance for the Janus CS3 covariate protocol.

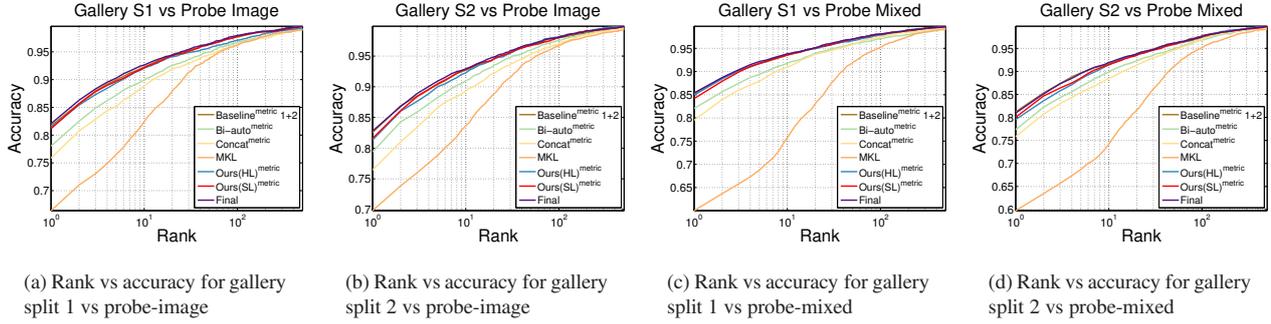

(a) Rank vs accuracy for gallery split 1 vs probe-image
(b) Rank vs accuracy for gallery split 2 vs probe-image
(c) Rank vs accuracy for gallery split 1 vs probe-mixed
(d) Rank vs accuracy for gallery split 2 vs probe-mixed

Figure 1. Rank vs accuracy for open set face identification performance. All results are reported after metric learning

|  | 1e-04 | 1e-03 | 1e-02 | 1e-01 |
|---|---|---|---|---|
| Overall | 0.4707 | 0.6639 | 0.8396 | 0.9488 |
| forehead visible = $-1$ | 0.4429 | 0.6300 | 0.8296 | 0.9506 |
| forehead visible = 0 | 0.3589 | 0.5243 | 0.7008 | 0.8827 |
| forehead visible = 1 | 0.4868 | 0.6860 | 0.8473 | 0.9485 |

Table 4. Covariate "forehead visible" verification performance for the Janus CS3 covariate protocol.

|  | 1e-04 | 1e-03 | 1e-02 | 1e-01 |
|---|---|---|---|---|
| Overall | 0.4707 | 0.6639 | 0.8396 | 0.9488 |
| gender = 0 | 0.3065 | 0.5384 | 0.7445 | 0.8904 |
| gender = 1 | 0.4741 | 0.6436 | 0.8206 | 0.9420 |

Table 5. Covariate "gender" verification performance for the Janus CS3 covariate protocol.

|  | 1e-04 | 1e-03 | 1e-02 | 1e-01 |
|---|---|---|---|---|
| Overall | 0.4707 | 0.6639 | 0.8396 | 0.9488 |
| indoor = $-1$ | 0.3508 | 0.5661 | 0.8052 | 0.9488 |
| indoor = 0 | 0.3884 | 0.5450 | 0.7114 | 0.8848 |
| indoor = 1 | 0.5188 | 0.7055 | 0.8552 | 0.9494 |

Table 6. Covariate "indoor" verification performance for the Janus CS3 covariate protocol.

|  | 1e-04 | 1e-03 | 1e-02 | 1e-01 |
|---|---|---|---|---|
| Overall | 0.4707 | 0.6639 | 0.8396 | 0.9488 |
| nose mouth visible = $-1$ | 0.3685 | 0.5718 | 0.8078 | 0.9486 |
| nose mouth visible = 0 | 0.3930 | 0.5502 | 0.7156 | 0.8793 |
| nose mouth visible = 1 | 0.5134 | 0.7033 | 0.8573 | 0.9494 |

Table 7. Covariate "nose mouth visible" verification performance for the Janus CS3 covariate protocol.

|  | 1e-04 | 1e-03 | 1e-02 | 1e-01 |
|---|---|---|---|---|
| Overall | 0.4707 | 0.6639 | 0.8396 | 0.9488 |
| skin tone = 1 | 0.6115 | 0.7665 | 0.8832 | 0.9556 |
| skin tone = 2 | 0.4298 | 0.5708 | 0.7144 | 0.8632 |
| skin tone = 3 | 0.1581 | 0.4099 | 0.7359 | 0.9299 |
| skin tone = 4 | 0.3513 | 0.4974 | 0.6699 | 0.8541 |
| skin tone = 5 | 0.2981 | 0.4062 | 0.5324 | 0.7446 |
| skin tone = 6 | 0.3890 | 0.5701 | 0.7215 | 0.8639 |

Table 8. Covariate "skin tone" verification performance for the Janus CS3 covariate protocol.

compared pair, the skin tone are of the same color for both images. It is noted that skin tone = 1 achieves the best performance over the other cases.

**Gender**: Gender = 1 means for the compared pair, the gender for the compared images is male. Gender = 0 means for the compared pair, the gender for the compared images is female.

**Indoor**: Indoor = $-1$ means for the compared pair, one image is taken indoor and the other one is taken outdoor. Indoor = 0 means for the compared pair, both images are taken outdoor and Indoor = 1 means that both image are taken indoor.

**Nose mouth visible**: Nose mouth visible = $-1$ means for the compared pair, nose mouth are visible in one image and invisible in the other one. Nose mouth visible = 0 means nose mouth are invisible for both images and nose mouth visible = 1 means nose mouth are visible for both images. Noted that nose mouth visible = 1 achieves the best performance over the other two cases.

**Skin tone**: Skin tone = $1, 2, 3, 4, 5$ or $6$ means for the

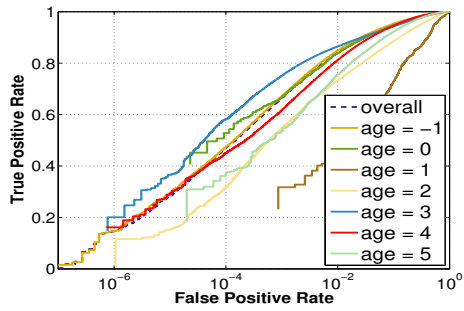
(a) age

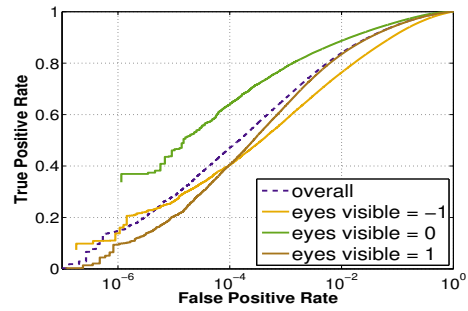
(b) eyes visible

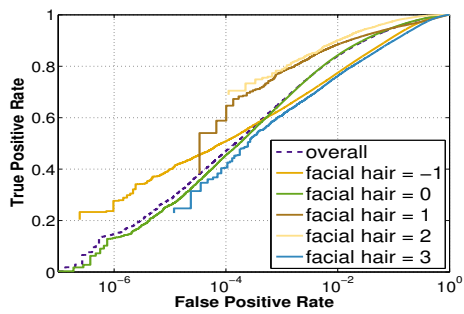
(c) facial hair

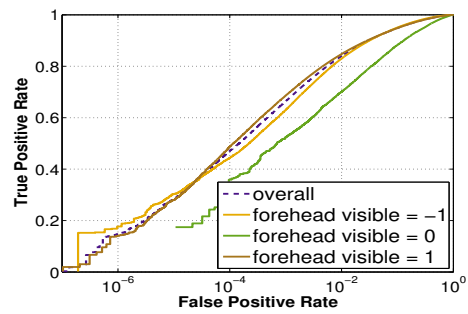
(d) forehead visible

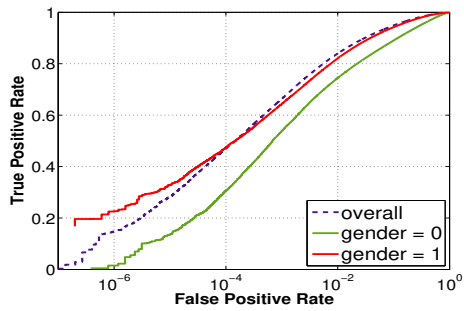
(e) gender

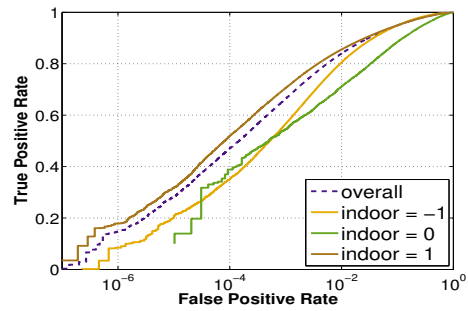
(f) indoor

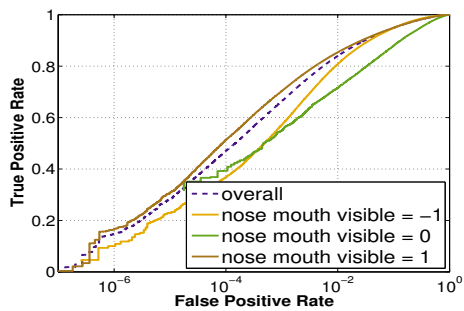
(g) nose mouth visible

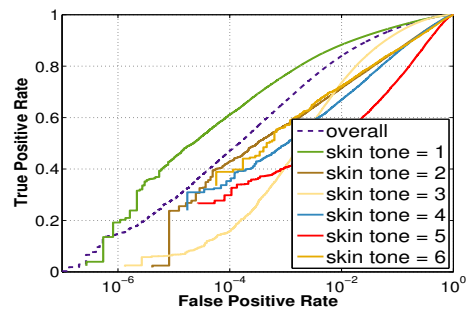
(h) skin tone

Figure 2. Face verification performance of eight different covariates from Janus CS3 covariate protocol